\documentclass[journal,twoside,web]{ieeecolor}
\usepackage{generic}
\usepackage{cite}
\usepackage{amsmath,amssymb,amsfonts}
\usepackage{algorithm}
\usepackage{algorithmic}
\usepackage{graphicx}
\usepackage{algorithm,algorithmic}
\usepackage{hyperref}
\usepackage{booktabs}
\usepackage{multirow}
\usepackage{makecell}
\hypersetup{hidelinks=true}
\usepackage{textcomp}
\def\BibTeX{{\rm B\kern-.05em{\sc i\kern-.025em b}\kern-.08em
    T\kern-.1667em\lower.7ex\hbox{E}\kern-.125emX}}
\begin{document}
\title{KAT-GNN: A Knowledge-Augmented Temporal Graph Neural Network for Risk Prediction in Electronic Health Records}

\author{Kun-Wei Lin, Yu-Chen Kuo, Hsin-Yao Wang and Yi-Ju Tseng \IEEEmembership{Member, IEEE}
\thanks{This study was supported by grants from the National Science and Technology Council, Taiwan (NSTC 114-2221-E-A49-061), the Higher Education Sprout Project of the National Yang Ming Chiao Tung University and MOE, Taiwan (CGMH-NYCU-114-CORPG2P0072), and Chang Gung Memorial Hospital (CMRPG2P0342).
}
\thanks{Kun-Wei Lin and Yu-Chen Kuo are with the Institute of Computer Science and Engineering, National Yang Ming Chiao Tung University, Hsinchu, Taiwan (e-mail: kwlin.cs12@nycu.edu.tw, yckuo.cs12@nycu.edu.tw).
Hsin-Yao Wang is with the School of Medicine, National Tsing Hua University, Hsinchu, Taiwan (e-mail: mdhsinyaowang@gmail.com).
Yi-Ju Tseng is with the Department of Computer Science, National Yang Ming Chiao Tung University, Hsinchu, Taiwan and Computational Health Informatics Program, Boston Children’s Hospital, Boston, MA, USA. (Corresponding author, e-mail: yjtseng@nycu.edu.tw). This work has been submitted to the IEEE for possible publication. Copyright may be transferred without notice, after which this version may no longer be accessible.
}}

\maketitle

\begin{abstract}
Clinical risk prediction using electronic health records (EHRs) is vital to facilitate timely interventions and clinical decision support. However, modeling heterogeneous and irregular temporal EHR data presents significant challenges. We propose \textbf{KAT-GNN} (Knowledge-Augmented Temporal Graph Neural Network), a graph-based framework that integrates clinical knowledge and temporal dynamics for risk prediction. KAT-GNN first constructs modality-specific patient graphs from EHRs. These graphs are then augmented using two knowledge sources: (1) ontology-driven edges derived from SNOMED CT and (2) co-occurrence priors extracted from EHRs. Subsequently, a time-aware transformer is employed to capture longitudinal dynamics from the graph-encoded patient representations. KAT-GNN is evaluated on three distinct datasets and tasks: coronary artery disease (CAD) prediction using the Chang Gung Research Database (CGRD) and in-hospital mortality prediction using the MIMIC-III and MIMIC-IV datasets. KAT-GNN achieves state-of-the-art performance in CAD prediction (AUROC: 0.9269 $\pm$ 0.0029) and demonstrated strong results in mortality prediction in MIMIC-III (AUROC: 0.9230 $\pm$ 0.0070) and MIMIC-IV (AUROC: 0.8849 $\pm$ 0.0089), consistently outperforming established baselines such as GRASP and RETAIN. Ablation studies confirm that both knowledge-based augmentation and the temporal modeling component are significant contributors to performance gains. These findings demonstrate that the integration of clinical knowledge into graph representations, coupled with a time-aware attention mechanism, provides an effective and generalizable approach for risk prediction across diverse clinical tasks and datasets.
\end{abstract}


\section{Introduction}
\label{sec:introduction}
\IEEEPARstart{R}{isk} prediction is a fundamental task in clinical decision support, facilitating early detection of patients at risk and enabling timely intervention~\cite{amirahmadi2023deep}. For example, predicting the future risk of coronary artery disease (CAD) allows clinicians to identify high-risk individuals in advance and initiate preventive strategies before severe cardiac events occur. Similarly, early prediction of in-hospital mortality in acute care settings enables timely allocation of intensive resources and informs clinical decision-making during critical periods~\cite{malakar2019review}. The increasing adoption of Electronic Health Records (EHRs) has generated large volumes of longitudinal patient data, creating new opportunities to develop comprehensive data-driven predictive models.

\begin{figure*}[t]
    \centering
    \includegraphics[width=\textwidth]{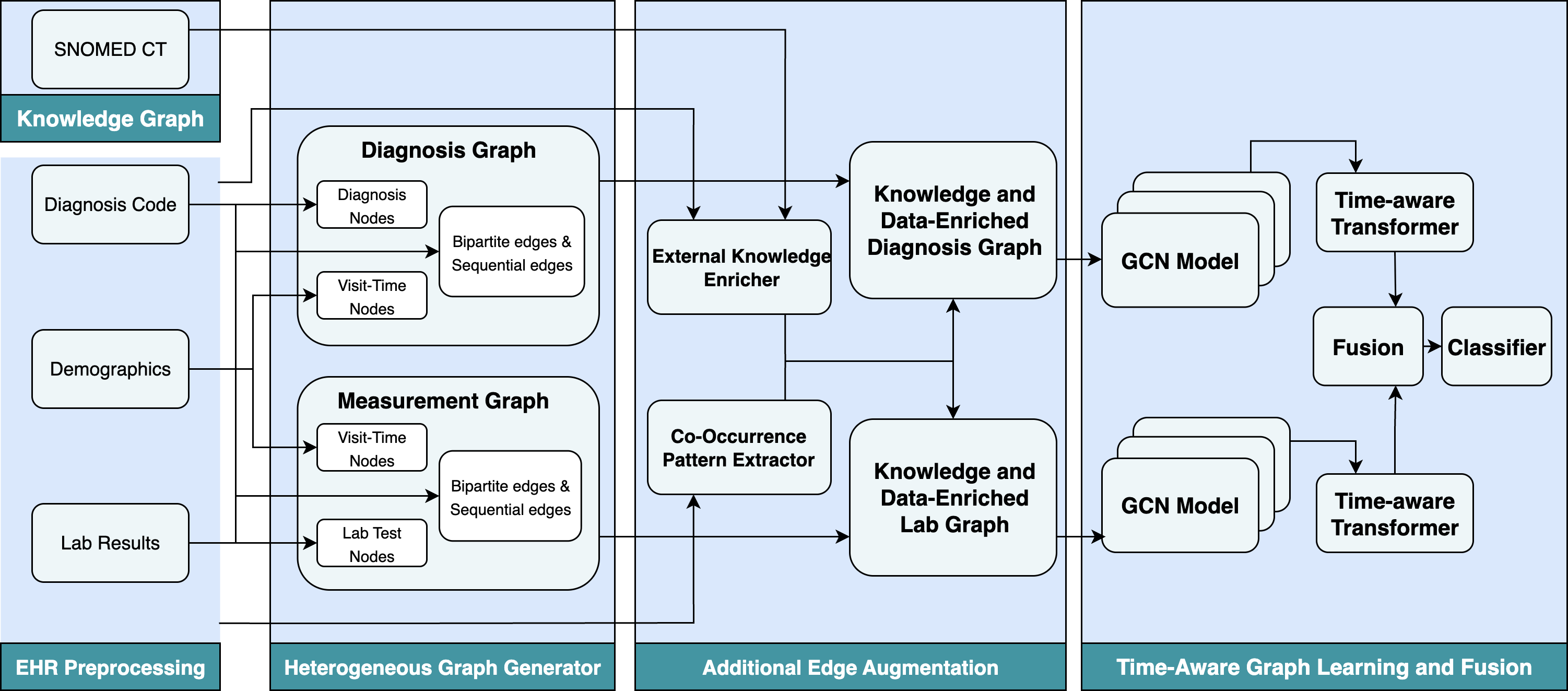}
    \caption[Overview of KAT-GNN framework]{\textbf{Overview of KAT-GNN framework.}
    The KAT-GNN framework comprises four stages: preprocessing, graph construction, edge augmentation, and time-aware graph learning and fusion.}
    \label{fig:pipeline}
\end{figure*}

Despite its potential, leveraging EHR data for risk prediction remains challenging~\cite{tayefi2021challenges}. First, heterogeneity arises from multiple modalities, including diagnoses, laboratory results, vital signs, and demographic information, which differ in structure and scale. Second, sparsity is common, since each encounter records only a limited subset of clinical measurements, resulting in fragmented and weakly connected representations~\cite{wu2010prediction}. Third, temporal dynamics are essential because disease progression and treatment response unfold over time and require models that capture longitudinal dependencies. Finally, effective knowledge integration remains a significant challenge. Many deep learning models neglect domain knowledge, such as medical ontologies~\cite{zhang2022graph} and co-occurrence patterns~\cite{lee2020harmonized}, which could enrich representations with clinically meaningful connections.

A substantial body of research has examined representation learning for EHR data. Early methods, such as DeepPatient~\cite{miotto2016deep} and DoctorAI~\cite{choi2016doctor}, leveraged unsupervised or recurrent neural networks to derive patient embeddings, but did not capture the relational structure among clinical entities. To address this gap, graph-based methods have emerged to explicitly encode dependencies among clinical entities. MedGCN~\cite{mao2022medgcn}, for example, constructs a heterogeneous graph over patients, encounters, labs, and medications, but it does not represent the temporal order between encounters of the same patient, thereby limiting longitudinal modeling. GCT~\cite{choi2020learning} introduces sequential links across visits, yet its hierarchical structure restricts direct cross-modality communication. For instance, a laboratory test can influence a diagnosis only through an intermediate treatment node, which increases propagation depth and hinders the learning of meaningful cross-modality patterns. To address the challenge of knowledge integration, knowledge-enriched models have also been explored. GSKN~\cite{zhang2022graph} integrates ontology-derived subgraphs, but its focus on local neighborhoods may overlook broader semantic relationships present in medical ontologies. HORDE~\cite{lee2020harmonized} leverages co-occurrence patterns to improve patient representations, but its reliance on unstructured clinical notes limits its applicability to structured EHR.

In summary, existing approaches face three major limitations: (i) inadequate modeling of temporal dependencies across visits, (ii) restricted cross-modality interactions arising from rigid graph hierarchies, and (iii) constrained or domain-specific strategies for incorporating external knowledge.

To address these challenges and limitations, we propose \textbf{KAT-GNN} (Knowledge-Augmented Temporal Graph Neural Network). This work makes the following main contributions:
\begin{enumerate}
    \item We propose a novel, graph-based framework that constructs modality-specific patient graphs (representing entities such as diagnoses and lab measurements) and employs a time-aware transformer to effectively capture longitudinal dependencies and patient health trajectories across visits.
    \item We introduce a dual-source knowledge augmentation strategy that integrates external, hierarchical knowledge from SNOMED CT with internal, statistical co-occurrence priors extracted from the EHRs. This approach effectively enriches graph representations with clinically meaningful connections.
    \item We provide a comprehensive evaluation of KAT-GNN on two critical prediction tasks (coronary artery disease and in-hospital mortality) using three large-scale datasets (CGRD, MIMIC-III, and MIMIC-IV \cite{johnson2023mimic}). Our results show that KAT-GNN achieves state-of-the-art performance, consistently outperforming established temporal and graph-based baselines.
\end{enumerate}

The source code is available at https://github.com/DHLab-TSENG/kat-gnn-ehr-risk

\section{Method}

\subsection{Framework Overview}
Figure~\ref{fig:pipeline} illustrates the KAT-GNN framework, which consists of four stages: \textbf{EHR preprocessing, heterogeneous graph generator, additional edge augmentation, and time-aware graph learning and fusion}.

In the \textbf{EHR preprocessing} stage, raw, structured clinical data (e.g., diagnoses, laboratory tests, demographics) are extracted, cleaned, and organized into modality-specific inputs for each patient.
Next, the \textbf{heterogeneous graph generator} stage uses these prepared inputs to build  modality-specific graphs. A \textit{diagnosis graph} is built to encode event-based disease information, and a \textit{measurement graph} is constructed to represent continuous clinical measurements, such as laboratory tests and vital signs. Both graphs incorporate bipartite edges linking clinical entities (nodes) to visits, and sequential edges capturing the temporal order of those visits.
As for \textbf{additional edge augmentation} stage, we enrich these base graphs by integrating additional edges from two sources of clinical knowledge: ontology-driven edges derived from SNOMED CT and empirical associations derived from co-occurrence statistics.
Finally, the \textbf{time-aware graph learning and fusion} stage learns the final patient representation. This stage involves three key steps: (1)encoding the node features within the knowledge-enriched graphs using graph convolutional networks, (2) applying time-aware attention mechanisms to capture both local and global visit dependencies, and (3) adaptively fusing multimodal embeddings for the downstream prediction task.

\subsection{Heterogeneous Graph Generator}

\begin{figure}[h]
    \centering
    \includegraphics[width=0.5\textwidth]{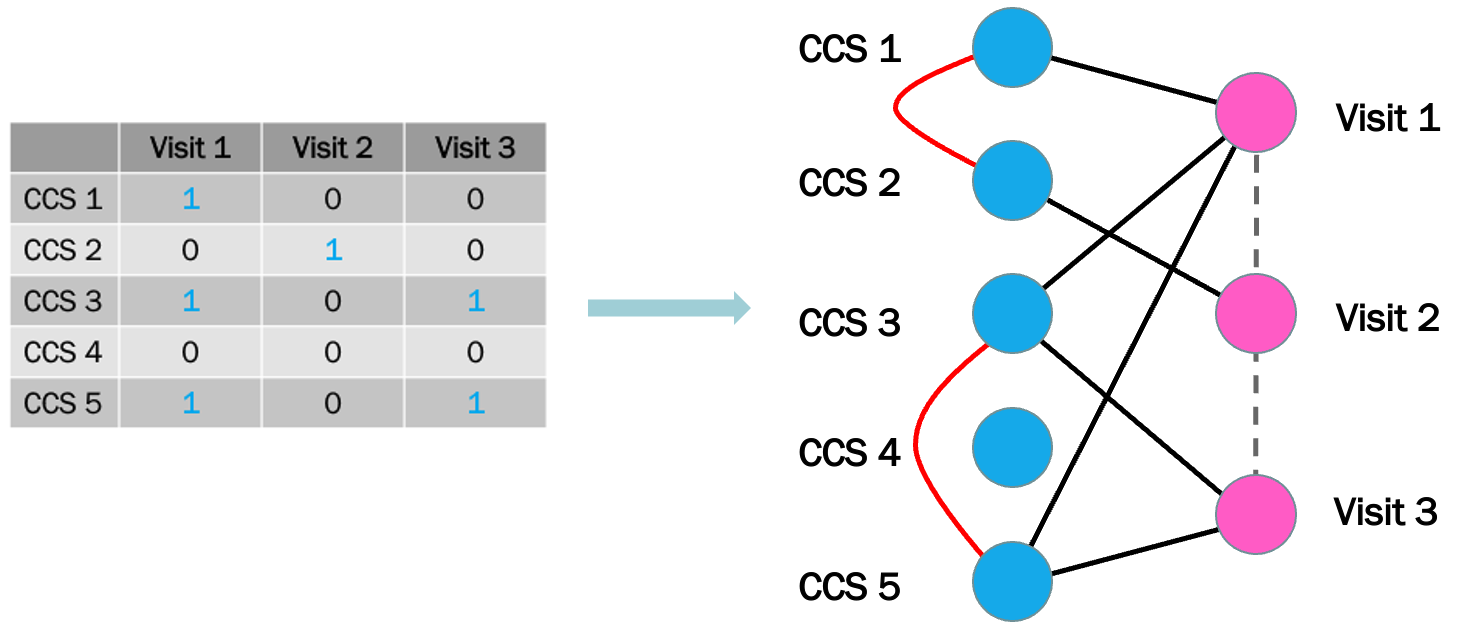}
    \caption{\textbf{Illustration of patient-specific graph construction.}
    The table on the left represents diagnosis records across multiple visits,
    where a value of 1 indicates that the patient was diagnosed with the corresponding CCS code during that visit,
    and 0 indicates absence.
    Each unique diagnosis and visit is represented as a node,
    with diagnosis nodes shown in \textcolor{blue}{blue} and visit nodes shown in \textcolor{magenta}{pink}.
    Black edges connect diagnoses to visits according to the EHR table,
    forming the fundamental bipartite structure of the graph. Gray dashed edges link consecutive visits to model the patient's temporal progression.
    Red edges indicate additional semantic connections introduced through knowledge augmentation,
    derived from external ontologies or co-occurrence statistics,
    linking clinically related diagnosis nodes to enrich the graph structure.}
    \label{fig:graph_construction}
\end{figure}

\subsubsection{Diagnosis Graph}
The diagnosis graph is designed to capture the categorical and temporal structure of diagnostic information. It is a bipartite graph containing two types of nodes: \textit{diagnosis nodes} and \textit{visit nodes} (Figure~\ref{fig:graph_construction}). Diagnosis nodes represent categories defined by the Clinical Classifications Software (CCS) ~\cite{elixhauser2021pl}. We use CCS because it abstracts the large number of International Classification of Diseases (ICD) codes into higher-level clinical categories, which not only mitigates data sparsity but also provides a unified representation across both ICD-9\cite{ccs_tool} and ICD-10\cite{ccsr_tool} codes present in the dataset. Visit nodes correspond to timestamps of patient encounters.

\textbf{Initial Node Features.}
Diagnosis nodes are represented as one-hot vectors in the diagnosis space, whereas visit-time nodes encode demographic attributes such as age and sex. The feature vector for a diagnosis node is defined as $\mathbf{v}_{\text{diag}} = [\mathbf{n}_{\text{diag}}, \mathbf{0}_{\text{demo}}]$, where $\mathbf{n}_{\text{diag}}$ is the one-hot diagnosis vector. The feature vector for a visit-time node is given by $\mathbf{v}_{\text{visit}} = [\mathbf{0}_{\text{diag}}, \mathbf{n}_{\text{demo}}]$,
where $\mathbf{n}_{\text{demo}}$ represents demographic attributes.

\textbf{Edges.}
The graph contains two types of edges to capture both diagnosis occurrence and chronology. Bipartite edges connect visit-time nodes to all diagnosis nodes recorded during the encounter. Sequential edges link consecutive visit-time nodes, explicitly modeling the temporal progression of the patient's record (Figure~\ref{fig:graph_construction}).

\subsubsection{Measurement Graph}

The measurement graph represents continuous clinical variables, such as laboratory tests and vital signs. A key challenge is integrating continuous values into a discrete graph structure. To address this, we discretized each measurement item (e.g., Serum Creatinine) into $B$ quantile-based bins (e.g., $B=4$), representing different value levels (e.g., Very Low, Low-Normal, High-Normal, Very High). Each of these item-bin pairs (e.g., Serum Creatinine - Very High) is modeled as a unique node in the graph. This discretization transforms continuous values into discrete entities that preserve clinical semantics. When both laboratory and vital sign data are available, they are modeled as separate measurement graphs.

\textbf{Initial Node Features.}
The total number of measurement nodes is $N_{\text{meas}} \times B$, where $N_{\text{meas}}$ is the number of unique measurement items.

The feature vector for a measurement node $m$ is defined as $\mathbf{v}_{\text{meas}} = [\mathbf{n}_{\text{meas}}, \mathbf{0}_{\text{demo}}]$, where $\mathbf{n}_{\text{meas}} \in \mathbb{R}^{N_{\text{meas}} \times B}$ is a one-hot vector identifying that specific item-bin combination, and $\mathbf{0}_{\text{demo}} \in \mathbb{R}^{N_{\text{demo}}}$ is a zero vector in the demographic space. Visit-time nodes follow the same design as in the diagnosis graph $\mathbf{v}_{\text{visit}} = [\mathbf{0}_{\text{meas}}, \mathbf{n}_{\text{demo}}]$
, where $\mathbf{0}_{\text{meas}} \in \mathbb{R}^{N_{\text{meas}} \times B}$ is a zero vector in the measurement space, and $\mathbf{n}_{\text{demo}} \in \mathbb{R}^{N_{\text{demo}}}$ encodes demographic attributes.

\textbf{Edges.}
Edges in the measurement graph are designed to capture both value-dependent associations and temporal continuity. Bipartite edges link each visit-time node to the specific measurement node corresponding to the values observed during that visit. Under the discretization scheme, the edge connects to the node representing the quantile bin of the observed value, thereby encoding measurement intensity. Finally, similar to the diagnosis graph, temporal edges are introduced.

\subsection{Edge Augmentation: Ontology-Driven Edges}
To mitigate data sparsity and enrich structural semantics, we first incorporate ontology-driven edges derived from SNOMED CT~\cite{lee2014literature, chang2021use}, a comprehensive clinical terminology encoding hierarchical relationships among medical concepts. Two complementary strategies are applied: diagnosis concept alignment and concept search for measurements.

\subsubsection{Diagnosis Concept Alignment}
Each CCS diagnosis category is aligned with SNOMED CT concepts through a multi-step mapping process. CCS categories are first expanded to ICD-10 codes using official mappings~\cite{ccs_icd10pcs}, which are then linked to SNOMED CT concepts via the I-MAGIC algorithm~\cite{imagic_mapper}.

The semantic relatedness between two CCS categories is then quantified by the depth of their lowest common subsumer (LCS) in the SNOMED CT hierarchy:
\begin{equation}
\text{dist}(\text{CCS}_i, \text{CCS}_j) =
    \frac{1}{|E_i||E_j|}\sum_{e_i \in E_i}\sum_{e_j \in E_j}\text{LCS}(e_i,e_j)
\end{equation}
where $E_i$ and $E_j$ denote the sets of mapped SNOMED CT concepts. Smaller distances indicate stronger semantic similarity between the corresponding CCS categories. To construct ontology-driven edges, we rank all possible node pairs by their distance values and select those with the smallest distances (i.e., the strongest semantic similarity) to form new connections. Since the total number of potential pairwise relations grows quadratically with the number of nodes (\(O(n^2)\)), we limit the augmentation to only the top percentage of the most similar pairs. This strategy effectively controls graph densification while ensuring that only the most semantically meaningful edges are introduced into the diagnosis graph.

\subsubsection{Measurement Concept Search}
Unlike diagnoses, laboratory tests and vital signs lack standardized mappings to SNOMED CT. We therefore adopt a keyword-based search strategy, inspired by~\cite{sung2023mapping}, to identify the corresponding SNOMED CT ontology concepts for measurement items. The search is implemented with the PyMedTermino2 library under the owlready2 ontology framework~\cite{lamy2015pymedtermino}.

\textbf{Preprocessing.} Each measurement label (lab test or vital sign) is standardized into a clinically precise query term. In the standardization process, abbreviations are expanded (e.g., ``WBC'' $\rightarrow$ ``white blood cell count''), contextual terms are added (e.g., ``hematocrit'' $\rightarrow$ ``measurement of hematocrit''), symbols are normalized (e.g., ``$\alpha$-fetoprotein'' $\rightarrow$ ``alpha fetoprotein''), and ambiguous phrases are clarified (e.g., ``O2 saturation pulseoxymetry'' $\rightarrow$ ``SpO2'', ``Glucose (AC)'' $\rightarrow$ ``glucose fasting'').

\textbf{Ontology Query and Filtering.} The processed terms are used to search for candidate SNOMED CT concepts. Candidate concepts are retained only if they belong to appropriate hierarchies in SNOMED CT ontology, including \textit{Procedure/Measurement} (concept ID: \texttt{128927009}, \texttt{122869004}) or \textit{Observable entity} (concept ID: \texttt{363787002}), and if their attributes indicate quantitative measurements (e.g., \texttt{129266000}, \texttt{129265001}, \texttt{30766002}).

\textbf{Redundancy Removal.} The ontology search may return multiple hierarchically related candidate concepts (e.g., a general concept and its specific sub-concepts). To ensure a single, consistent mapping, we retain only the most general concept—defined as the ancestor of all other candidates in the hierarchy—is retained. This links each measurement item to a single, semantically broad representative, resulting in a one-to-one mapping.

\textbf{Semantic Distance.} Pairwise similarity between measurement items is quantified by the depth of their lowest common subsumer (LCS) in the SNOMED CT hierarchy:
\begin{equation}
    \text{dist}(\text{Meas}_i, \text{Meas}_j) = \text{LCS}(e_i, e_j)
\end{equation}
where $e_i$ and $e_j$ denote the SNOMED CT concepts associated with $\text{Meas}_i$ and $\text{Meas}_j$, respectively. These distances define ontology-driven edges between measurement nodes.


\subsection{Edge Augmentation: Co-Occurrence-Driven Edges}
While ontology-based edges incorporate curated medical knowledge, co-occurrence-driven edges capture empirical dependencies directly from the data. These edges reflect cohort-specific patterns such as diagnoses or measurements that frequently appear together in a patient visit, thereby complementing the graph structure with data-driven statistical associations.

\subsubsection{Support and Lift}
Each patient visit is treated as a transaction containing a set of clinical concepts. For any pair of items $A$ and $B$, the \textbf{support} is defined as
\begin{equation}
\text{Support}(A, B) = \frac{\#\{\text{visits containing } A,B\}}{\#\{\text{total visits}\}}
\end{equation}
and the \textbf{lift} is computed as
\begin{equation}
\text{Lift}(A, B) = \frac{\text{Support}(A, B)}{\text{Support}(A)\cdot \text{Support}(B)}
\end{equation}
A lift greater than one indicates that the pair co-occurs more often than expected by chance.

\subsubsection{Edge Construction}
Unweighted edges are added between measurement nodes and diagnosis nodes, respectively, with $\text{Lift}(A,B) > 1$, indicating a positive association. These co-occurrence edges enhance the graph by capturing cohort-specific statistical relationships, thereby enabling the model to leverage both external ontologies and intrinsic data patterns.

\subsection{Time-Aware Graph Learning and Fusion}

\subsubsection{Graph Encoders}
Each modality-specific graph (diagnosis and measurement) is independently encoded using a multi-layer graph neural network. We consider three encoder variants: Graph Convolutional Networks (GCN)~\cite{kipf2016semi}, Graph Attention Networks (GAT)~\cite{velickovic2017graph}, and GATv2~\cite{brody2021attentive}. Formally, given a graph $G=(\mathcal{V},\mathcal{E})$ with node features $\mathbf{X}$, the layer-wise update in GCN is
\begin{equation}
    \mathbf{H}^{(l+1)} = \sigma \left( \hat{\mathbf{D}}^{-\tfrac{1}{2}} \hat{\mathbf{A}} \hat{\mathbf{D}}^{-\tfrac{1}{2}} \mathbf{H}^{(l)} \mathbf{W}^{(l)} \right),
\end{equation}
where $\hat{\mathbf{A}}$ is the adjacency matrix with self-loops, $\hat{\mathbf{D}}$ its degree matrix, and $\sigma(\cdot)$ a non-linearity.
GAT and GATv2 replace the normalized aggregation with learnable attention coefficients, enabling context-dependent neighbor weighting. These encoders produce node-level embeddings $\mathbf{Z}^{(v)}$ for each modality $v$.

\subsubsection{Local and Global Time-Aware Attention}

To capture temporal dynamics across visits, we propose two complementary attention mechanisms adapted from HiTaNet~\cite{luo2020hitanet}. This design ensures that the model captures both short-term, recency-based dependencies (via \textit{local attention}) and long-range temporal progression throughout the full patient trajectory (via \textit{global attention}).

\paragraph{Local Attention.}
Given a sequence of visit-time node embeddings $\{\mathbf{x}_i\}_{i=1}^N$ and their corresponding time intervals to the prediction index date $\{\tau_i\}_{i=1}^N$, we construct two complementary inputs: the \textit{visit-time node embeddings} derived from the graph encoder and the \textit{temporal embeddings} obtained by transforming the time intervals.
Each temporal embedding is generated through a non-linear projection:
\begin{align}
    \mathbf{t}_i &= 1 - \tanh\left(\mathbf{W}_1 \cdot \tau_i^2 \right), \\
    \mathbf{e}_i &= \text{concat}\big(\mathbf{x}_i, \mathbf{W}_2 \cdot \mathbf{t}_i\big)
\end{align}
where $\mathbf{W}_1 \in \mathbb{R}^{1 \times 64}$ and $\mathbf{W}_2 \in \mathbb{R}^{64 \times d}$ are learnable parameters.

A scalar attention score $s_i$ is then computed:
\begin{equation}
    s_i = \mathbf{w}^\top \mathbf{e}_i, \quad \alpha_i = \frac{\exp(s_i / \sqrt{d})}{\sum_j \exp(s_j / \sqrt{d})}
\end{equation}
where $\mathbf{w} \in \mathbb{R}^{2d}$ is a learnable vector.

The local representation is obtained as a weighted sum of visit embeddings:
\begin{equation}
    \mathbf{z}_{\text{local}} = \sum_{i=1}^N \alpha_i \mathbf{x}_i
\end{equation}

\paragraph{Global Attention.}
To capture long-range dependencies, we first compute a global query vector from the average visit embedding:
\begin{align}
    \mathbf{h} &= \frac{1}{N} \sum_{i=1}^N \mathbf{x}_i, \\
    \mathbf{q} &= \text{ReLU}\left(\mathbf{W}_q \cdot \mathbf{h}\right)
\end{align}
where $\mathbf{W}_q \in \mathbb{R}^{d \times d}$ is a trainable projection matrix.

Each temporal embedding $\mathbf{t}_i$ is then scored against $\mathbf{q}$ using scaled dot-product attention:
\begin{equation}
    \beta_i = \frac{\exp\big(\mathbf{q}^\top \mathbf{t}_i / \sqrt{d}\big)}{\sum_j \exp\big(\mathbf{q}^\top \mathbf{t}_j / \sqrt{d}\big)}
\end{equation}

The global representation is defined as:
\begin{equation}
    \mathbf{z}_{\text{global}} = \sum_{i=1}^N \beta_i \mathbf{t}_i
\end{equation}

\begin{figure}[h]
    \centering
    \includegraphics[width=1.05\linewidth]{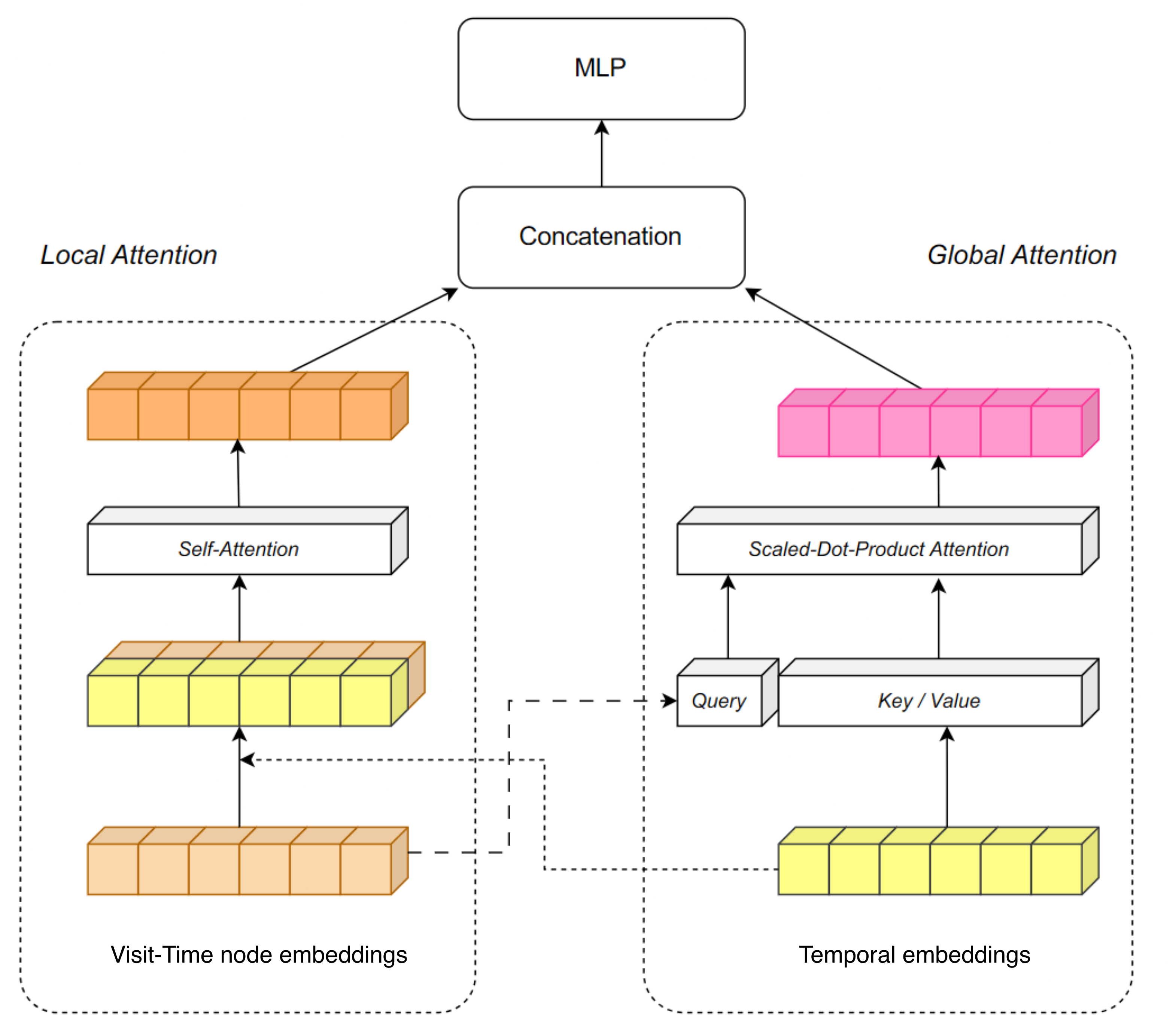}
    \caption{\textbf{Architecture of local and global time-aware attention.}
The module processes two distinct inputs: (1) the \emph{visit-time node embeddings}, which are derived from visit-time nodes in the graph, representing temporal positions of visits, and (2) the \emph{temporal embeddings}, which are generated from time intervals to the index date.}
    \label{fig:time_aware_attention}
\end{figure}

\paragraph{Final Representation.}
The two modules are combined to yield a comprehensive temporal representation:
\begin{equation}
    \mathbf{z}_{\text{time}} = \mathbf{z}_{\text{local}} + \mathbf{z}_{\text{global}}
\end{equation}
This joint design ensures that the model captures both short-term visit-specific effects and long-term temporal progression patterns (Fig.~\ref{fig:time_aware_attention}).

\subsubsection{Adaptive Fusion}
To combine heterogeneous modalities, we employ an adaptive late fusion mechanism. Let $\mathbf{z}_{\text{time}}^{(v)}$ denote the embedding from modality $v$. A learnable weight vector $\boldsymbol{\alpha}$ is normalized by softmax:
\begin{equation}
\tilde{\alpha}_v = \frac{\exp(\alpha_v)}{\sum_{v'} \exp(\alpha_{v'})},
\end{equation}
and the fused representation is
\begin{equation}
\mathbf{z}_{\text{fused}} = \sum_{v} \tilde{\alpha}_v \cdot \mathbf{z}_{\text{time}}^{(v)}.
\end{equation}
This design enables the model to adaptively prioritize modalities according to their task relevance. A fully connected prediction head then uses $\mathbf{z}_{\text{fused}}$ to produce the final risk prediction.

\section{Experimental Setting}

\begin{table*}[t]
    \centering
    \caption{\textbf{Model Performance on CGRD, MIMIC-III, and MIMIC-IV.}
    Results are reported as mean (standard deviation). Statistical significance is denoted by *, **, and *** for comparisons against the best-performing model at $p<0.05$, $p<0.01$, and $p<0.001$, respectively.}
    \label{tab:model_comparison}
    \begin{tabular}{lcccccc}
    \toprule
    \multirow{2}{*}{\textbf{Model}} &
    \multicolumn{2}{c}{\textbf{CGRD}} &
    \multicolumn{2}{c}{\textbf{MIMIC-III}} &
    \multicolumn{2}{c}{\textbf{MIMIC-IV}} \\
    \cmidrule(lr){2-3} \cmidrule(lr){4-5} \cmidrule(lr){6-7}
     & AUROC & AUPRC & AUROC & AUPRC & AUROC & AUPRC \\
    \midrule
    Transformer   & 0.8773 (0.0055)*** & 0.5837 (0.0171)*** & 0.8888 (0.0076)*** & 0.5632 (0.0253)*** & 0.8146 (0.0097)*** & 0.4664 (0.0226)*** \\
    RETAIN        & 0.8859 (0.0052)*** & 0.6185 (0.0116)*** & 0.9128 (0.0047)*** & 0.6427 (0.0207)*** & 0.8528 (0.0058)*** & 0.5935 (0.0140)*** \\
    StageNet      & 0.8856 (0.0078)*** & 0.5646 (0.0180)*** & 0.9020 (0.0062)*** & 0.6340 (0.0225)*** & 0.8474 (0.0077)*** & 0.6104 (0.0180)*** \\
    GRASP         & 0.8891 (0.0038)*** & 0.6274 (0.0134)*** & 0.9052 (0.0055)*** & 0.6181 (0.0215)*** & 0.8478 (0.0063)*** & 0.5780 (0.0146)*** \\
    TRANS         & 0.7487 (0.0337)*** & 0.3219 (0.0310)*** & 0.7687 (0.0400)*** & 0.3484 (0.0446) *** & 0.7277 (0.0317)*** & 0.3387 (0.0402)*** \\
    \midrule
    \textbf{KAT-GNN (GCN)}   & \textbf{0.9269 (0.0029)} & \textbf{0.6764 (0.0118)} & \textbf{0.9230 (0.0070)} & \textbf{0.6719 (0.0220)} & 0.8820 (0.0069) & 0.6520 (0.0156)*** \\
    \textbf{KAT-GNN (GAT)}   & 0.9206 (0.0055)*** & 0.6695 (0.0096)* & 0.9201 (0.0075) & 0.6716 (0.0229) & \textbf{0.8849 (0.0089)} & \textbf{0.6714 (0.0175)} \\
    \textbf{KAT-GNN (GATv2)} & 0.9201 (0.0069)*** & 0.6716 (0.0142) & 0.9166 (0.0077)*** & 0.6645 (0.0225) & 0.8828 (0.0079) & 0.6657 (0.0145) \\
    \bottomrule
    \end{tabular}
\end{table*}

\subsection{Datasets}
\textbf{Private Dataset (CGRD).} We use the Chang Gung Research Database (CGRD)~\cite{tsai2017chang}, which contains de-identified EHR collected between 2001 and 2018, including diagnoses, laboratory tests, and demographics. The study cohort consists of patients undergoing routine cardiac check-ups~\cite{liu2023temporal, lin2024graph}. The protocol was approved by the Institutional Review Board of the Chang Gung Medical Foundation (IRB No. 201801771B0).

\textbf{Public Datasets (MIMIC-III and MIMIC-IV).} We further evaluate on the publicly available MIMIC-III (v1.4)~\cite{johnson2016mimic} and MIMIC-IV (v2.2)~\cite{johnson_mimic-iv_2023}, which include ICU admissions between 2001–2012 and 2008–2019 at Beth Israel Deaconess Medical Center, respectively. Both datasets contain diagnoses, laboratory tests, vital signs, and demographics. Access was granted under the PhysioNet credentialed user agreement~\cite{johnson2023mimicivphysionet}.

\subsection{Prediction Tasks}
\textbf{CAD Prediction (CGRD).} The task is to predict the future onset of coronary artery disease (CAD) for patients from the CGRD cardiac check-up cohort. Cases are defined as patients with $\geq 2$ CAD-related diagnoses within 30–365 days. The index date is the first CAD diagnosis. Controls without CAD are indexed at their last recorded visit~\cite{liu2023temporal, lin2024graph}. The final cohort consisted of 53,567 patients, of whom 6,272 (11.7\%) were identified as positive cases. For model input, we extract EHR data from the five years preceding a defined index date while leaving a one-year gap between the end of this history and the prediction target. This design ensures that the model forecasts future CAD risk without accessing information too close to the outcome, thereby preventing data leakage. For feature preprocessing, all diagnoses (ICD-9/10) are mapped to CCS categories\cite{ccsr_tool, ccs_tool} to ensure semantic consistency, resulting in 276 codes. Laboratory items recorded in at least 20\% of patients are retained, resulting in 80 laboratory features.

\textbf{In-Hospital Mortality (MIMIC-III/IV).} The objective is to predict patient-level in-hospital mortality after admission to the ICU. For patients with multiple ICU stays, only the first ICU stay from the last hospital admission is retained. This selection process resulted in a final cohort of 17,250 patients for MIMIC-III (mortality rate: 12.5\%) and 21,107 patients for MIMIC-IV (mortality rate: 16.1\%). The index date for each patient is defined as the admission date of that final hospitalization, and the prediction target is the mortality status during that admission. For model input, diagnoses are collected prior to the admission, while lab tests and vitals are taken from the first 48 hours of the ICU stay. For feature preprocessing, diagnoses (ICD-9/10) are mapped to CCS categories\cite{ccsr_tool, ccs_tool}. Lab results are extracted from the first 48 hours of ICU admission; the same 20\% threshold is applied, resulting in 64 laboratory features. Vital signs are standardized by name and unit.

\subsection{Baselines}
We compare our framework against several established models for clinical prediction, implemented using PyHealth~\cite{yang2023pyhealth}:
\begin{itemize}
    \item \textbf{Transformer}~\cite{vaswani2017attention}: applies multi-head self-attention over visit sequences.
    \item \textbf{RETAIN}~\cite{choi2016retain}: a sequential model with dual-level attention for interpretability.
    \item \textbf{StageNet}~\cite{gao2020stagenet}: models stage transitions with stage-aware LSTMs and temporal convolutions.
    \item \textbf{GRASP}~\cite{zhang2021grasp}: augments patient representations by incorporating similar-patient cohorts via graph convolution.
    \item \textbf{TRANS}~\cite{chen2024predictive}: a heterogeneous graph transformer that encodes both structural and temporal relations among medical events.
\end{itemize}

\subsection{Implementation Details}
All models are implemented in PyTorch (v1.12.1) with PyTorch Geometric (v2.3.1). Graph-based modules are initialized with Xavier uniform initialization. Models are trained with the Adam optimizer (learning rate = 0.01, weight decay = $10^{-6}$, batch size = 128). A dropout rate of 0.5 and the OneCycleLR scheduler~\cite{smith2019super} are applied. Experiments are conducted on a single NVIDIA RTX 4090 GPU.

\subsection{Training and Evaluation Protocol}
Datasets are split into 64\%/16\%/20\% for training, validation, and testing. Each experiment is repeated 30 times with different random seeds. Model selection is based on the highest validation Area Under the Precision-Recall Curve (AUPRC). Performance is reported using Area Under the ROC Curve (AUROC) and AUPRC.

The ablation studies were conducted on the CGRD dataset, which offers broader temporal coverage and a substantially larger patient cohort, enabling more representative analysis of model components. In contrast, the MIMIC-III and MIMIC-IV datasets were utilized primarily as public benchmarks to facilitate fair comparison with existing approaches.

\subsection{Statistical Analysis}
To account for training variability, the results are reported as mean $\pm$ standard deviation on 30 evaluation runs. Statistical significance between models is evaluated using paired t-tests, with Bonferroni correction for multiple comparisons. Significance levels are denoted as * ($p<0.05$), ** ($p<0.01$), and *** ($p<0.001$). Analyses are performed with \texttt{scipy.stats}.

\section{Results}

\subsection{Overall Performance Comparison}

We evaluated the proposed framework on two prediction tasks: future onset of CAD  in CGRD and in-hospital mortality in MIMIC-III and MIMIC-IV (Table~\ref{tab:model_comparison}). Performance was measured by AUROC and AUPRC, which provide complementary perspectives in the presence of class imbalance\cite{mcdermott2024closer}.

In the CAD prediction task using the CGRD dataset, the GCN-based variant of KAT-GNN achieved state-of-the-art performance (AUROC: 0.9269; AUPRC: 0.6764), followed closely by the GAT and GATv2 variants. All three variants of KAT-GNN substantially outperformed sequential baselines such as RETAIN and StageNet, as well as patient-similarity and heterogeneous graph approaches, including GRASP and TRANS.

In the in-hospital mortality prediction task in MIMIC-III, KAT-GNN with the GCN variant achieved the best overall performance (AUROC: 0.9230; AUPRC: 0.6719), with GAT and GATv2 also remaining competitive. All variants with our proposed models consistently outperformed baselines, demonstrating robustness in a public cohort that differs from CGRD in temporal structure and coding characteristics.

In MIMIC-IV, KAT-GNN with the GAT variant achieved the best performance (AUROC: 0.8849; AUPRC: 0.6714), followed closely by GATv2 and GCN variants.
This suggests that the GAT's adaptive attention mechanism may be particularly effective for modeling the complex interactions in this acute care setting, though all KAT-GNN variants consistently outperformed the baselines.

The KAT-GNN configurations used in this comparison were finalized based on ablation studies detailed in the later session. The optimal settings—which balance semantic enrichment with graph sparsity—were found to be: 10 quantile bins for continuous variable discretization, and an edge augmentation threshold corresponding to the top 3\% of ontology-driven edges for diagnosis graphs and 5\% for measurement graphs.

Overall, KAT-GNN consistently and significantly outperforms all baseline models across three distinct datasets and two prediction tasks (Table~\ref{tab:model_comparison}). The consistent improvements across both chronic disease risk prediction (CGRD) and acute care mortality prediction (MIMIC-III/IV) confirm the effectiveness and generalizability of the proposed knowledge-augmented temporal graph framework.

\subsection{Effect of Discretization Granularity}

Table~\ref{tab:bin_ablation} evaluates the impact of discretizing laboratory values into different numbers of bins. This ablation was only conducted on the measurement graph derived from the CGRD dataset, ensuring consistency by isolating a single modality. Performance improves steadily from 1 (AUROC: 0.9198) to 10 (AUROC: 0.9227) bins, indicating that incorporating value-level information enhances the expressiveness of measurement nodes. The best results were obtained with 10 bins, suggesting that moderate granularity provides an effective balance between detail and robustness. Beyond this point, as seen with 25 bins (AUROC: 0.9213), performance slightly declines, likely due to sparsity introduced by excessive fragmentation. These findings confirm that fine-grained but not overly detailed discretization yields the most stable predictive gains.

\begin{table}[h]
\centering
\caption{\textbf{Effect of lab value discretization granularity.}
Results are reported as mean (standard deviation). Statistical significance is denoted by *, **, and *** for comparisons against the best-performing model.}
\label{tab:bin_ablation}
\begin{tabular}{lcc}
\toprule
\textbf{Bins} & AUROC & AUPRC  \\
\midrule
1   & 0.9198 (0.0031)\textsuperscript{***} & 0.6461 (0.0138)\textsuperscript{***}  \\
2   & 0.9214 (0.0041) & 0.6567 (0.0190)\textsuperscript{**}  \\
4   & 0.9224 (0.0037) & 0.6652 (0.0158)  \\
10  & \textbf{0.9227} (0.0024) & \textbf{0.6710} (0.0082)  \\
25  & 0.9213 (0.0026)\textsuperscript{***} & 0.6673 (0.0079)\textsuperscript{*}  \\
\bottomrule
\end{tabular}
\end{table}

\subsection{Influence of Ontology-Driven Edges}

Table~\ref{tab:kg_edge_ablation} evaluates the effect of integrating ontology-driven edges derived from SNOMED CT.

For the diagnosis graph, we observe a clear benefit from knowledge enrichment. Performance (AUROC) improves from 0.7564 at 0\% augmentation to a peak of 0.7649 when the top 3\% edges are added. However, adding too many edges is detrimental; performance declines to 0.7526 at 10\% augmentations. This suggests that moderate semantic enrichment enhances predictive ability, while excessive augmentation introduces redundancy that disrupts the patient-specific structure.

In contrast, the benefit of the measurement graph (using CGRD laboratory data) is limited. AUROC shows only a slight increase, rising from 0.9227 at the 0\% baseline to a peak of 0.9244 at 5\% augmentation. Meanwhile, AUPRC actually reaches its maximum at the 0\% baseline (0.6710) and does not improve with any added edges.

These results suggest that ontology-based enrichment is more effective for the sparser, higher-level diagnosis information. The laboratory measurement nodes, which already encode granular value-level information (via discretization), appear to provide strong discriminative value, yielding limited benefit from external priors.

\begin{table}[h]
\centering
\caption{\textbf{Effect of adding knowledge graph edges.}
Statistical significance is denoted by *, **, and *** for comparisons against the 0\% baseline.}
\label{tab:kg_edge_ablation}

\begin{tabular}{c|cc}
\toprule
\% KG Edges& AUROC& AUPRC\\\midrule
 \multicolumn{3}{c}{Diagnosis Graph}\\
\midrule
0\%  & 0.7564 (0.0078) & 0.3070 (0.0119) \\
1\%  & 0.7598 (0.0057)\textsuperscript{*} & 0.3172 (0.0105)\textsuperscript{***} \\
2\%  & 0.7628 (0.0084)\textsuperscript{***} & 0.3221 (0.0140)\textsuperscript{***} \\
3\%  & \textbf{0.7649} (0.0069)\textsuperscript{***} & \textbf{0.3244} (0.0119)\textsuperscript{***} \\
4\%  & 0.7604 (0.0082)\textsuperscript{**} & 0.3191 (0.0117)\textsuperscript{***} \\
5\%  & 0.7612 (0.0064)\textsuperscript{**} & 0.3185 (0.0102)\textsuperscript{***} \\
10\% & 0.7526 (0.0124) & 0.3107 (0.0133) \\
\midrule
 \multicolumn{3}{c}{Measurement Graph}\\
\midrule
 0\%  & 0.9227 (0.0024) &\textbf{0.6710} (0.0082) \\
 1\%  & 0.9240 (0.0026)\textsuperscript{*} &0.6618 (0.0111)\textsuperscript{***} \\
 2\%  & 0.9239 (0.0027) &0.6624 (0.0106)\textsuperscript{**} \\
 3\%  & 0.9238 (0.0026)\textsuperscript{**} &0.6620 (0.0093)\textsuperscript{***} \\
 4\%  & 0.9243 (0.0022)\textsuperscript{***} &0.6638 (0.0093)\textsuperscript{***} \\
 5\%  & \textbf{0.9244} (0.0021)\textsuperscript{***} &0.6652 (0.0085)\textsuperscript{**} \\
 10\% & 0.9243 (0.0024)\textsuperscript{***} &0.6652 (0.0078)\textsuperscript{**} \\
 \bottomrule
\end{tabular}

\end{table}

\subsection{Validation of Ontology-Driven Edges}

Table~\ref{tab:kg_random_ablation} compares ontology-driven edges against randomly added edges of equal quantity. This ablation is conducted separately on two single-modality graphs: the diagnosis graph and the laboratory-based measurement graph from the CGRD dataset. For the diagnosis graph, ontology-derived edges are integrated at 3\%, yielding statistically significant improvements in both AUROC and AUPRC compared with the random baseline. For the measurement graph, ontology edges are added at 5\%, where improvements are smaller but remain significant in AUPRC. These results confirm that the observed performance gains originate from semantically meaningful relations rather than from increased graph density alone.


\begin{table}[h]
\centering
\caption{\textbf{Ontology-driven vs. random edges.}}
\label{tab:kg_random_ablation}

\begin{tabular}{l|cc}
\toprule
& AUROC & AUPRC \\
\midrule
 \multicolumn{3}{c}{Diagnosis Graph}\\
\midrule
With ontology edges & \textbf{0.7649}\textsuperscript{***} (0.0069) & \textbf{0.3244}\textsuperscript{*} (0.0119) \\
With random edges   & 0.7594 (0.0089) & 0.3202 (0.0143) \\
\midrule
 \multicolumn{3}{c}{Measurement Graph}\\
 \midrule
 With ontology edges & \textbf{0.9244} (0.0021) &\textbf{0.6652}\textsuperscript{**} (0.0085) \\
 With random edges   & 0.9234 (0.0045) &0.6550 (0.0201) \\
 \bottomrule
\end{tabular}

\end{table}

\subsection{Influence of Co-Occurrence Edges}

Table~\ref{tab:cooccurrence_ablation} reports the effect of adding co-occurrence edges derived from empirical frequency patterns in the EHR. This ablation is conducted on two single-modality graphs constructed from the CGRD dataset: a diagnosis graph and a laboratory-based measurement graph. Incorporating these edges produces consistent and statistically significant gains in both diagnosis and measurement graphs. For the diagnosis graph, AUROC increases from 0.7564 to 0.7604 ($p<0.01$) and AUPRC from 0.3070 to 0.3156 ($p<0.001$). For the measurement graph, AUROC improves from 0.9227 to 0.9242 ($p<0.001$) and AUPRC from 0.6710 to 0.6735 ($p<0.05$). These findings indicate that co-occurrence priors enhance graph connectivity and provide predictive signal beyond that captured by EHR-derived edges alone.


\begin{table}[h]
\centering
\caption{\textbf{Effect of co-occurrence edges.}}
\label{tab:cooccurrence_ablation}

\begin{tabular}{l|cc}
\toprule
& AUROC & AUPRC \\
\midrule
 \multicolumn{3}{c}{Diagnosis Graph}\\
\midrule
Without co-occurrence & 0.7564 (0.0078) & 0.3070 (0.0119) \\
With co-occurrence    & \textbf{0.7604}\textsuperscript{**} (0.0075) & \textbf{0.3156}\textsuperscript{***} (0.0134) \\
\midrule
 \multicolumn{3}{c}{Measurement Graph}\\
 \midrule
 Without co-occurrence & 0.9227 (0.0024) &0.6710 (0.0082) \\
 With co-occurrence    & \textbf{0.9242}\textsuperscript{***} (0.0025) &\textbf{0.6735}\textsuperscript{*} (0.0090) \\
\bottomrule

\end{tabular}
\end{table}

\subsection{Contribution of Time-Aware Attention Mechanisms}

Table~\ref{tab:timeaware_ablation} evaluates the impact of incorporating time-aware attention mechanisms into the GCN backbone of KAT-GNN. This experiment employs the dual-graph configuration, which integrates both the diagnosis and laboratory measurement graphs from the CGRD dataset. The model uses the best-performing settings identified in previous ablations: 3\% ontology edges for the diagnosis graph, 5\% for the measurement graph, 10-bin discretization for laboratory values, and inclusion of co-occurrence edges in both graphs. The temporal module captures progression patterns across visits by weighting clinical events according to their relative timing.

Adding the time-aware transformer leads to statistically significant gains, with AUROC improving from 0.9184 to 0.9269 ($p<0.001$) and AUPRC from 0.6618 to 0.6764 ($p<0.001$). These results demonstrate that explicitly modeling temporal dynamics enhances the predictive capacity of graph-based representations, highlighting the importance of temporal context in clinical risk prediction.

\begin{table}[h]
\centering
\caption{\textbf{Effect of time-aware transformer.}}
\label{tab:timeaware_ablation}
\begin{tabular}{lcc}
\toprule
Model & AUROC & AUPRC \\
\midrule
GCN (without time aware)& 0.9184 (0.0033) & 0.6618 (0.0085) \\
GCN (with time aware)& \textbf{0.9269}\textsuperscript{***} (0.0029) & \textbf{0.6764}\textsuperscript{***} (0.0118) \\
\bottomrule
\end{tabular}
\end{table}

\section{Discussion}
\label{sec:discussion}

In this study, we proposed a patient-specific graph construction framework that integrates heterogeneous modalities in EHRs and incorporates temporal dynamics. The design is modality-aware and accommodates diverse data types, including event-based diagnoses, irregular laboratory measurements, and high-frequency vital signs. By explicitly modeling patient trajectories as graphs, the framework enhances the structural representation of clinical records.

Our findings align with prior work, such as GSKN~\cite{zhang2022graph}, which demonstrated the benefit of integrating external knowledge graphs for predictive modeling. Ontology-driven augmentation with SNOMED CT edges consistently improved performance, although the gains were sensitive to the proportion of added edges. For diagnosis graphs, performance peaked at 3\% edge integration, whereas excessive augmentation (e.g., 10\%) reduced performance due to over-densification. In contrast, laboratory graphs showed only modest improvements, with the best AUROC observed at 5\% integration. This asymmetry underscores the need to balance semantic enrichment with graph sparsity in order to preserve informative patient-specific structures.

Compared to ontology-based augmentation, co-occurrence-driven edges offered a more conservative yet consistently beneficial strategy. By restricting edges to statistically meaningful associations (lift $>$ 1), the resulting graphs retained cohort-specific dependencies without excessive densification. These results emphasize the dual importance of semantic quality and structural sparsity in knowledge-aware graph modeling.

Temporal information is a critical determinant of clinical outcomes. Previous studies, such as HiTaNet~\cite{luo2020hitanet}, have demonstrated the importance of temporal context by modeling visit sequences through recurrent or index-based representations. Building on this insight, KAT-GNN extends temporal modeling by explicitly incorporating visit-to-index time intervals within a time-aware transformer module. Ablation results show that this design significantly improves predictive performance, underscoring the complementary role of temporal dynamics and graph-based structural representations in capturing disease progression.

Finally, comparisons against strong baselines—including Transformer~\cite{vaswani2017attention}, RETAIN~\cite{choi2016retain}, StageNet~\cite{gao2020stagenet}, GRASP~\cite{zhang2021grasp}, and TRANS~\cite{chen2024predictive}—demonstrated consistent superiority of our framework across chronic disease (in CGRD) and critical care (in MIMIC III and IV) tasks. These results underscore the generalizability of graph-temporal modeling for diverse clinical prediction settings.

Several limitations warrant discussion. First, ontology integration for laboratory tests is constrained by the lack of a standardized approach for mapping to SNOMED CT. The keyword-based mapping approach, adapted from prior work~\cite{sung2023mapping}, is susceptible to semantic noise, which may partly account for the inconsistent AUPRC improvements reported in Table~\ref{tab:kg_edge_ablation}. Future work could address this limitation through the adoption of standardized coding systems or ontology alignment methods to improve reliability. Second, we maintained separate modality-specific graphs rather than constructing a unified multimodal graph. While integration could capture richer cross-modality interactions, it would substantially increase input dimensionality and edge density, raising risks of sparsity, overparameterization, and computational overhead. Future work could explore scalable multimodal graph architectures to balance representational richness with efficiency.

\section{Conclusion}
\label{sec:conclusion}

This work presents KAT-GNN, a graph-based framework for clinical risk prediction that unifies dual-source knowledge (ontology-driven enrichment and data-driven co-occurrence) with time-aware temporal modeling. By constructing modality-specific graphs and incorporating both external and data-driven relations, the framework effectively captures semantic, statistical, and temporal dimensions of patient health trajectories.

Extensive experiments on three large-scale datasets (CGRD, MIMIC-III, MIMIC-IV) demonstrated that KAT-GNN consistently outperforms transformer-, recurrent-, and graph-based baselines on both chronic disease and acute care prediction tasks. The results highlight the value of structured graph representations enriched with knowledge and temporal dynamics to advance predictive modeling in healthcare.

\section*{References}

\bibliographystyle{IEEEtran}
\bibliography{ref}

\end{document}